\documentclass{article}
\usepackage[final]{tackling_climate_workshop_style}
\usepackage[utf8]{inputenc}
\usepackage[T1]{fontenc}
\usepackage{hyperref}
\usepackage{url}
\usepackage{booktabs}
\usepackage{amsmath}
\usepackage{graphicx}
\usepackage{xcolor}
\usepackage{caption}
\usepackage{float}   
\hypersetup{colorlinks=true, linkcolor=black, citecolor=black, urlcolor=black}
\title{WildfireSpreadBench: The Metric Decides the Model\\
in Wildfire Spread Prediction}

\author{%
Arin Gopakumar \\
UC Berkeley \\
\texttt{aringopakumar@berkeley.edu}
\And
Marco Pannozzo \\
Purdue University \\
\texttt{mpannozz@purdue.edu}
}

\begin{document}
\maketitle
\begin{abstract}
Machine learning is being increasingly used to predict where active wildfires will burn the
following day, helping inform evacuation boundaries and containment lines. Most models
are evaluated using Average Precision (AP), which summarizes performance across all
decision thresholds, although acting on a forecast requires choosing one. We
benchmarked five discriminative architectures and one generative model on
WildfireSpreadTS using a shared evaluation pipeline and two input configurations. We
found that model rankings varied depending on whether performance was measured by AP or
by threshold-dependent metrics like F1 and IoU. At the default 0.5 threshold, the highest-AP model flagged 4 to 5 times the observed active-fire area and ranked fifth of six on F1 and IoU, and the most recall-heavy model 
flagged 16 to 23 times. The three models whose flagged area stayed close to the observed area had AP scores 24 to 37\% lower.
Across models, we identified three distinct prediction profiles: over-predicting, balanced, 
and under-predicting, which AP alone could not distinguish. Expanding the input from 7 to 23 channels changed AP by
0.03 on average, against a 0.21 to 0.24 spread across models. Ultimately, these results show that
AP alone can favor models whose predictions are poorly suited for operational wildfire
forecasting.
\end{abstract}
\section{Introduction}
Fire seasons in the western United States have grown longer and more severe, largely due to anthropogenic warming \citep{abatzoglou2016}, while smoke exposure has increased alongside them \citep{burke2021}. Where and how a fire burns today shapes where it burns tomorrow, and incident teams have to anticipate that to advise on evacuation and containment lines.
Progress in technology has made this job easier; physics-based rate-of-spread models
\citep{sullivan2009} now sit alongside models trained on satellite data
\citep{huot2022}, and WildfireSpreadTS \citep{gerard2023} gave the field a
shared benchmark. Convolutional, recurrent and attention-based architectures have been
benchmarked on it \citep{lahrichi2026}, and generative ones on simulated fire data \citep{yu2026}.
The majority are scored with AP, which is defensible since it needs no threshold, and precision--recall beats ROC-AUC when the positive class is rare \citep{davis2006,saito2015}. However, deployment is different since fire teams draw outlines on a map, which means picking a threshold. If AP and threshold metrics disagree on model performance, evaluating solely on AP may favor the wrong one for field use. We tested six models and contribute:
\begin{itemize}\setlength{\itemsep}{0pt}
\item \textbf{A unified benchmark} of all six models under one metric implementation.
\item \textbf{Evidence that metrics change rankings}, with AP and F1 at 0.5 picking different winners.
\item \textbf{A reporting standard} pairing AP with thresholded metrics and predicted-to-observed area.
\end{itemize}
\section{Related Work}\label{sec:Related work}
\textbf{Datasets and multi-temporal learning.} The advancement and rigorous evaluation of 
spatio-temporal architectures depend entirely on the availability of standardized, 
continuous data. Early foundational benchmarks, such as the Next Day Wildfire Spread dataset,
catalyzed initial deep learning research by providing high-resolution static
snapshots of historical fires alongside environmental variables
\citep{huot2022}. While useful, such static transitions miss the continuous,
sequential evolution of a spreading fire. The WildfireSpreadTS dataset addresses 
this critical limitation by providing contiguous, 24-hour multi-modal time-series 
observations \citep{gerard2023}. Rather than a cross-sectional snapshot of the fire, 
WildfireSpreadTS captures 23 input data channels, including meteorological factors, 
land coverage, and ground truth active fire area across multi-day windows. This lets researchers evaluate architectures on their capacity to learn complex 
temporal dynamics and sequential physical interactions rather than pattern matching alone \citep{gerard2023}.

\textbf{Standardized evaluation metrics for imbalanced domains.} A major hurdle in wildfire spread 
prediction training and evaluation is extreme class imbalance. Because wildfire datasets consist of very 
few ``active fire'' pixels in comparison to ``no fire'' pixels, designing a loss function and
evaluation metric that captures the accuracy of predicting fire spread can be
challenging \citep{andrianarivony2024}. A model could achieve roughly 99.9\% accuracy
simply by predicting that no fire will occur anywhere. Consequently, sound benchmarking requires specialized metrics designed for imbalanced semantic segmentation. 
Intersection over Union (IoU) and the F1-score measure spatial overlap
directly, heavily penalizing models that over-predict the fire class. 
To further analyze a model's operational performance, Precision (the proportion of predicted fire pixels with an active-fire detection the next day) and Recall (the proportion of next-day active-fire pixels predicted by the model) are reported separately. For instance, in the context of emergency management, maximizing Recall is often
prioritized to prevent the catastrophic under-prediction of a fire's leading edge
\citep{rosch2024}. Finally, Average Precision summarizes the
precision--recall curve across all operational thresholds
\citep{gerard2023}. The gap between such threshold-free summaries and behavior at a chosen cutoff is a documented pitfall in biomedical image analysis \citep{maierhein2024, reinke2024pitfalls}. Weather forecast verification reports frequency bias and the critical success index for the same reason \citep{roebber2009visualizing}, which are our predicted-to-observed area ratio and IoU.

\textbf{Warning decisions and disaster response.}
Acting on a forecast is far different from evaluating it from a machine learning perspective. It means choosing a cutoff for warning and evacuation, and the hazard-warning literature treats that choice as central. Warning triggers link a predicted condition to a protective action for a target group, and their errors are false positives, where recommendations are stronger than the hazard warrants, and false negatives, where warnings are too weak \citep{cova2017warning}. Wildfire incident managers make such decisions under uncertainty about fire behavior, data quality, and the values at risk, and decision-support research aims to match tools to those uncertainties \citep{thompson2011uncertainty}. Social, institutional, and psychological factors also shape incident decisions \citep{thompson2014social}. Benchmark reporting rarely shows how a spread forecast behaves once a cutoff is applied, which is the behavior these frameworks act on.

\textbf{The gap.}
WildfireSpreadTS benchmarks report AP as the headline metric \citep{gerard2023, lahrichi2026}, and to our knowledge none checks whether the AP ranking holds once a threshold is applied. This study tests that, scoring established baselines alongside a custom BCE U-Net and an adapted Flow Matching model under one set of metrics. Because warning and incident decisions are made at a cutoff, we evaluate models by how they would behave operationally (flagged area relative to observed active-fire area, precision, and recall at a fixed threshold) alongside AP.

\section{Methods}
\textbf{Data.} WildfireSpreadTS \citep{gerard2023} comprises 13{,}607 daily
images across 607 U.S.\ fire events from January 2018 to October 2021, at
375\,m resolution over the 23 channels described in Section \ref{sec:Related work}. We write day $t$ for the most recent observed day of a fire and day $t{+}1$ for the day after, which is the day whose active fire mask every model predicts. ConvLSTM and UTAE receive days $t{-}4$ through $t$, whereas the other four receive day $t$ alone. All models are trained on 2018 and 2019 and tested on the 2021 season. The four dataset baselines also validate on 2020. To separate the effect of the model from the effect of the input data, we evaluate two configurations. Vegetation uses seven channels of reflectance, vegetation indices, and active-fire data, while All uses 23 channels.

\textbf{Models.} Five are discriminative, meaning they focus on predicting labels. Logistic Regression, a pixel-wise linear baseline; ResNet18 U-Net, the encoder--decoder released with the dataset
\citep{gerard2023,ronneberger2015,he2016}; ConvLSTM \citep{shi2015}, which models
day-to-day dependence recurrently; UTAE \citep{garnot2021}, a U-Net with a
temporal attention encoder for satellite time series; and our own BCE U-Net, a
segmentation U-Net trained with a positive-weighted binary cross-entropy
objective. Flow Matching \citep{lipman2023,liu2023} learns a continuous-time velocity field carrying noise to the data distribution and integrates it as an ODE at inference, so its prediction is sampled rather than thresholded from one forward pass. We include this modeling approach because fire spread is
stochastic, and the authors of the dataset expected label-noise-tolerant methods to aid in this regard \citep{gerard2023}.
\section{Results}
\begin{table}[H]
\centering\small
\setlength{\tabcolsep}{5.28pt}
\caption{The models predict the wildfire spread 24h ahead based on two input feature sets: \emph{Vegetation} and \emph{All}. The performance displayed is the test set AP, F1, IoU, Precision, and Recall for the 2021 test year, with threshold metrics computed at 0.5. Rows are grouped by operating-point profile, and bold marks the best.}
\vspace{4pt}
\label{tab:main}
\begin{tabular}{l cc cc cc cc cc}
\toprule
& \multicolumn{2}{c}{AP} & \multicolumn{2}{c}{F1} & \multicolumn{2}{c}{IoU}
& \multicolumn{2}{c}{P} & \multicolumn{2}{c}{R}\\
\cmidrule(lr){2-3}\cmidrule(lr){4-5}\cmidrule(lr){6-7}\cmidrule(lr){8-9}\cmidrule(lr){10-11}
Model & Veg & All & Veg & All & Veg & All & Veg & All & Veg & All\\
\midrule
BCE U-Net           & \textbf{0.562} & \textbf{0.529} & 0.308 & 0.329 & 0.182 & 0.197 & 0.186 & 0.203 & 0.896 & 0.861\\
UTAE                & 0.383 & 0.322 & 0.115 & 0.080 & 0.061 & 0.042 & 0.061 & 0.042 & \textbf{0.969} & \textbf{0.964}\\
ResNet18 U-Net      & 0.375 & 0.401 & 0.568 & \textbf{0.585} & 0.396 & \textbf{0.413} & 0.558 & \textbf{0.599} & 0.578 & 0.572\\
ConvLSTM            & 0.403 & 0.390 & \textbf{0.583} & 0.570 & \textbf{0.411} & 0.398 & \textbf{0.565} & 0.543 & 0.602 & 0.599\\
Logistic Regression & 0.352 & 0.371 & 0.548 & 0.566 & 0.378 & 0.394 & 0.501 & 0.551 & 0.606 & 0.581\\
Flow Matching       & 0.322 & 0.350 & 0.402 & 0.425 & 0.252 & 0.270 & 0.448 & 0.465 & 0.365 & 0.392\\
\bottomrule
\end{tabular}
\end{table}
\begin{figure}[H]
\centering
\includegraphics[width=\textwidth]{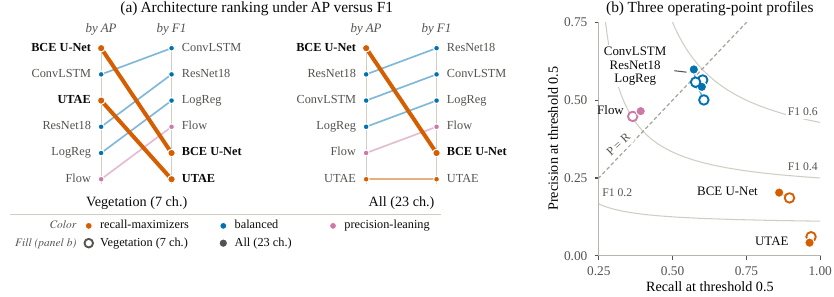}
\caption{\textbf{(a)} Model rank under AP against rank under F1, for both
feature sets; bolded lines mark ranks that move three places or more.
\textbf{(b)} The same twelve runs as precision--recall operating points, with
iso-F1 contours.}
\label{fig:main}
\end{figure}
\begin{figure}[H]
\centering
\includegraphics[width=\textwidth,height=100pt,keepaspectratio]{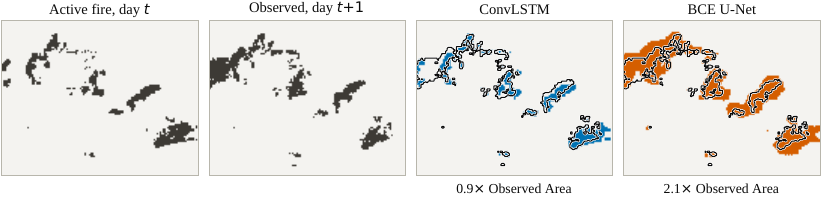}
\caption{Active fire on day $t$, the observed fire on
day $t{+}1$, and the ConvLSTM and BCE U-Net predictions at a 0.9 threshold, with the observed
footprint outlined on both prediction panels. BCE U-Net covers 2.1$\times$ the
observed area and merges the observed patches into roughly a dozen blobs; ConvLSTM covers 0.9$\times$.}
\label{fig:qual}
\end{figure}
\section{Discussion and Climate Impact}
\label{sec:discussion}
 
\textbf{Metric analysis.} AP is the standard metric in
machine learning for class-imbalanced datasets \citep{davis2006,saito2015}.
However, it may be over-relied upon in the context of wildfire spread. This is
demonstrated most strongly in this study by the BCE U-Net model, which earned the
highest AP score of all models across both feature sets. Despite its high AP, at the default 0.5 threshold it flags 4 to 5 times the observed active-fire area, and Figure~\ref{fig:qual} shows over-prediction even at 0.9. Its positive class weight moves the decision boundary away from 0.5 by construction, so the over-prediction reflects the operating point and not necessarily forecasting ability. AP cannot show it, because AP does not describe behavior at any one operating point \citep{maierhein2024}. \citet{rosch2024}
argue over-prediction is the more tolerable failure, since false positives can be
revised. At 16 to 23 times the observed active-fire area, which is what UTAE flags at 0.5, that tolerance runs out.
 
\textbf{Dataset and training choices.} Wildfire training datasets have been a focus
of the field for the past half-decade, from Next Day Wildfire Spread
\citep{huot2022} to WildfireSpreadTS and its expansion, WildfireSpreadTS+. A key finding affirmed by this study is that higher-complexity datasets
do not appear to be the immediate answer to improving machine learning applications in wildfire spread prediction. Where \citet{gerard2023} found that the full channel set reduced AP for
their temporal models, we found the seven vegetation channels gave metrics very similar to the full set. This,
combined with the WildfireSpreadTS+ finding that four further years of training data
did not improve accuracy \citep{lahrichi2026}, suggests that
raw data volume is not the primary bottleneck in wildfire machine learning. In contrast, modeling and training choices lead to large changes in output characteristics. The clearest one is how the loss handles class imbalance \citep{lin2017}. The two models trained with positive-weighted BCE, UTAE at a weight of 964 and BCE U-Net at 50, are the two over-predictors, while the three trained with unweighted Dice or Jaccard losses sit near the diagonal in Figure~\ref{fig:main}(b). Loss and architecture vary together in this benchmark, so we do not attribute the profiles to architecture alone. At the default 0.5 threshold, ConvLSTM as trained here reaches nine or more times the precision of UTAE. This degree of variability is not present when tailoring feature sets or gathering more input variables, indicating that near-term work should prioritize the training objective, calibration and threshold selection over adding input channels.
 
\textbf{Climate impact.} As anthropogenic warming intensifies fire seasons \citep{abatzoglou2016,burke2021}, effective climate adaptation requires deployable machine learning. However, at its default threshold, the highest-AP model here flags 4 to 5 times the observed active-fire area. Unneeded evacuation is costly, as documented for hurricanes \citep{whitehead2003million}, and repeated false alarms lower trust in later warnings, the cry wolf effect \citep{leclerc2015cry}.

\textbf{Reporting standard.} We recommend that wildfire spread benchmarks report AP together with F1, precision and recall at a threshold selected on a validation year under a stated rule, such as maximum F1 or a minimum recall. They should also report the ratio of predicted to observed active-fire area and the same metrics for the persistence forecast. An agency with a fixed response budget can instead fix the flagged area and compare recall across models, which makes the trade-off between false alarms and missed fire explicit.

\paragraph{Limitations.}
Our evaluation protocol limits how far the rankings can be read as evidence about operational suitability. First, every threshold-dependent metric is computed at a fixed threshold of 0.5, although the models differ in loss, class weighting, and score distribution. We chose a fixed threshold to emulate deployment, where a fire prediction is binary, and as a proof of concept that predictions can look drastically different at similar AP scores. However, a positive weight shifts the decision boundary away from 0.5 by construction, and the two most recall-heavy models are exactly the two trained with positive-weighted BCE (weights of 964 and 50), so part of the differences in F1, precision, recall, and flagged area reflects calibration and threshold choice rather than forecasting ability. Threshold calibration should therefore be implemented for deployment. The area ratios in Table~\ref{tab:appendix-area} describe behavior at the default threshold, not an intrinsic property of an architecture. We did not evaluate at validation-selected thresholds or at matched predicted-area budgets, which is the most important follow-up. Second, each configuration was run once, and the Flow Matching row is a single unseeded draw, so we report no confidence intervals and small rank differences should be treated with caution. Third, BCE U-Net and Flow Matching were developed without a validation set and with 2021 test metrics visible, so their rows are upper bounds and not clean held-out estimates. Finally, AP is estimated on a fixed grid of 200 thresholds, which drifts low for scores concentrated in a narrow band, as Flow Matching's are. Our results also cover a narrow slice of the problem. All numbers come from fold 0 of the twelve-fold protocol, which holds out 2021, the easiest of the four years by the dataset's own persistence measure (AP 0.287 against a four-year mean of 0.193) \citep{gerard2023}, so they describe relative behavior on this benchmark and not performance in other years, geographies, or fire regimes. Our labels are VIIRS active-fire detections, not final burned areas or operational perimeters, and they carry satellite false positives and missed detections, so ``observed area'' is an estimate of true active wildfire. 
\section{Conclusion}
We present WildfireSpreadBench, an evaluation study of six models for next-day wildfire spread prediction under a shared protocol, designed to test whether Average Precision and thresholded metrics select the same model. Discriminative models lead on every metric in both feature configurations, with BCE U-Net taking AP, while ConvLSTM and ResNet18 U-Net take F1, IoU, and precision. The generative model never leads, and ResNet18 U-Net beats it on all five metrics in both configurations; the label-noise tolerance that motivated its inclusion does not appear in these results. At the 0.5 threshold, Flow Matching is the only model that under-predicts, covering 0.8 times the observed active-fire area, where BCE U-Net covers 4 to 5 times. This contrast highlights an important difference between overall predictive performance and the practical behavior of the resulting fire maps. Which metric is reported, and at what threshold, therefore matters far more than the feature set. These results show that model selection can change substantially depending on whether evaluation emphasizes ranking quality or thresholded spatial predictions. Future work should widen the evaluation rather than the input. This calls for validation-selected thresholds, multiple seeds and multiple held-out years on varied fire data, while newer generative formulations should receive the same treatment, and the prioritized metrics should be validated against firefighters' judgment in real-world working settings. Code and configurations are available at \url{https://github.com/aringopakumar/WildfireSpreadBench}, covering all six models, the shared data pipeline, and the evaluation routine behind every number in Table~\ref{tab:main}.

\bibliographystyle{plainnat}
\bibliography{refs}


\appendix

\section{Dataset}\label{app:data}

\textbf{Composition.} WildfireSpreadTS \citep{gerard2023} contains 13{,}607
daily images across 607 wildfire events in the contiguous United States,
spanning January 2018 to October 2021 at 375\,m ground resolution. Events were
selected from GlobFire as fires larger than 1000 hectares, with four buffer days
added before and after each event, and each is converted to one HDF5 bundle of
shape $[T, 23, H, W]$. Image sizes vary between $304 \times 207$ and
$356 \times 308$ because of how Google Earth Engine partitions its output, so
$H$ and $W$ differ between fires. The four years are unevenly represented with 176 events
being from 2018, 74 from 2019, 201 from 2020 and 156 from 2021.

\textbf{Channel groups.} The 23 channels described in
Section~\ref{sec:Related work} come from seven source products and fall into the
eight groups in Table~\ref{tab:appendix-channels}. Active fire is derived from
the VIIRS 375\,m active fire product, surface reflectance from VNP09GA bands I1,
I2 and M11, the vegetation indices from VNP13A1, observed weather and drought
from GRIDMET, forecast weather from the Global Forecast System, land cover from
the MODIS yearly product under the IGBP classification, and topography from NASA
SRTM. Every other source is resampled bilinearly to the 375\,m resolution of the
active fire maps and aggregated into 24-hour windows beginning at midnight.

\begin{table}[h]
\centering\small
\caption{The 23 released channels by group, following the feature groupings used
in the ablation studies of \citet{gerard2023}, and which groups the
\emph{Vegetation} configuration draws on.}
\label{tab:appendix-channels}
\vspace{4pt}
\begin{tabular}{llcc}
\toprule
Group & Source product & Channels & In \emph{Vegetation}\\
\midrule
Active fire         & VIIRS VNP14IMG & 1 & Yes\\
Surface reflectance & VIIRS VNP09GA  & 3 & Yes\\
Vegetation indices  & VIIRS VNP13A1  & 2 & Yes\\
Weather             & GRIDMET        & 6 & No\\
ERC and drought     & GRIDMET        & 2 & No\\
Weather forecast    & GFS            & 5 & No\\
Topography          & SRTM           & 3 & No\\
Land cover          & MODIS MCD12Q1  & 1 & No\\
\midrule
Total               &                & 23 & \\
\bottomrule
\end{tabular}
\end{table}

The weather group holds minimum and maximum surface temperature, total
precipitation, wind speed, wind direction and specific humidity. The energy
release component and the Palmer Drought Severity Index come from the same
product but are grouped separately, following the ablation studies in
\citet{gerard2023}. The forecast group mirrors the weather group with two
exceptions: the Global Forecast System supplies mean temperature in place of a
minimum and a maximum, and it reports wind as speed and direction after
conversion from its native u and v components. Topography holds elevation
together with the slope and aspect derived from it.

\textbf{From 23 channels to 40 model inputs.} The models do not read the
released channels directly. Preprocessing expands the categorical land cover
channel into a one-hot encoding over the 17 IGBP classes and appends the
binarized active fire mask after standardization, so the tensor reaching the
model carries 40 channels rather than 23. The published parameter count for the
logistic regression baseline confirms this width independently, since that model
is a single $3 \times 3$ convolution and its 361 parameters are exactly
$40 \times 9 + 1$ \citep{gerard2023}. Under our \emph{Vegetation} configuration
the same model has 64 parameters.

Continuous channels are standardized with means and standard deviations
estimated on the training years alone. Two groups are held out of that
standardization. The angular channels, meaning wind direction, aspect and
forecast wind direction, are mapped through a sine transform instead, since a
linear rescaling of a quantity that wraps at 360 degrees carries no meaning. The
land cover class is left categorical for the one-hot expansion. The statistics
are fixed constants selected by the declared training years, so no test-year
statistic can reach training.

\textbf{Feature configurations.} In the released channel order the reflectance
and vegetation index channels occupy indices 0 to 4 and the active fire channel
occupies index 22, which after the one-hot expansion of land cover lands at
index 38, with the appended binary mask at 39. \emph{Vegetation} therefore keeps
processed indices $[0,1,2,3,4,38,39]$, corresponding to five reflectance and vegetation-index 
channels and two active-fire channels. \emph{All} keeps every
processed channel. Both settings follow \citet{gerard2023}, who report a
\emph{Vegetation} and an \emph{All} feature set and always include the fire masks
rather than counting them as a feature group. The two numeric labels used in the
body count different quantities. The 7 of \emph{Vegetation} is a model input
width, since the active fire channel enters twice, once standardized and once
binarized, and it draws on 6 released channels. The 23 of \emph{All} is a
released channel count, which the same preprocessing widens to 40 model inputs.

\textbf{Split protocol.} We use fold 0 of the cross-validation protocol defined
by the dataset authors. Training uses the 2018 and 2019 fire seasons, validation
uses 2020 for the models that use a validation set, and testing uses the
held-out 2021 season. Standardization statistics
come from 2018 and 2019 only, matching the training years. The full protocol in \citet{gerard2023} runs all twelve permutations of years across train, validation and test, which they describe as necessary given how much the
yearly distributions differ. Fold 0 also holds out the easiest of the four years
by their own persistence measure. Every number in Table~\ref{tab:main} comes from
fold 0, so the comparisons are paired across architectures, not
averaged over folds.

\textbf{Crops and augmentation.} Training and validation draw random
128-by-128-pixel crops. The crop window favors regions containing fire pixels
instead of being placed uniformly at random, and horizontal flips, vertical
flips and 90-degree rotations are applied as augmentation, with the angular
channels rotated to match so that wind direction and aspect stay physically
consistent with the transformed image.

\textbf{Test-time cropping.} Test evaluation does not use 128-by-128 crops. Each
scene is center-cropped to the nearest multiple of 32 in each dimension, which is
what the U-Net encoders require, and scored at batch size 1 because scene sizes
differ between fires. ConvLSTM, the only model that needs a fixed input size, runs tiled inference
across the scene. On the final row and column the crop window is aligned to the
bottom and right edges and overlapping predictions are overwritten, and the
tiles are aggregated into a full-resolution map before any metric is computed.

\textbf{Test-set alignment.} The number of samples a fire contributes depends on
how many leading days a model consumes, so the test set would otherwise differ
between the one-day and five-day models. The dataset class exposes a test
adjustment that skips the first $n_{\text{adj}} - n_{\text{leading}}$ samples of
each fire. Setting it to five throughout gives a skip of four for the one-day
models and zero for the five-day models, so all six are scored on an identical
set of targets beginning on the sixth day of each fire.

\textbf{Class imbalance.} About 0.1\% of pixels in the dataset carry an active
fire detection, and the rest do not \citep{gerard2023}. At that prevalence a
model predicting no fire anywhere reaches roughly 99.9\% accuracy, which is why
Section~\ref{sec:Related work} treats accuracy as uninformative here and why
Table~\ref{tab:main} reports precision and recall separately. It also bears on
how the AP column should be read, since AP is prevalence-dependent and the base
rate, not anything about the models, sets its absolute scale.

\textbf{Known issues in the released pipeline.} Two are worth recording, because
both affect any work built on this benchmark. The dataset class contained a bug
that the authors found after publication and corrected in the repository, and
they report that the corrected version gives slightly higher performance while
leaving the trends unchanged \citep{wsts_repo}. Separately, in February 2026
the authors noted that the angular channels are transformed through sine only,
where sine and cosine together would be needed to preserve direction, so wind
direction and aspect reach the model with some information lost
\citep{wsts_repo}. Both issues apply to every model we evaluate, so they do
not favor any architecture, but the second one falls entirely inside the
\emph{All} configuration and gives one concrete reason why the additional
channels may deliver less than they should.

\section{Models and Training}\label{app:models}

One harness evaluates all six architectures. Every model exposes a prediction
function returning a per-pixel score in $[0,1]$, and that function is handed to a
single evaluation routine that computes AP, F1, IoU, precision and recall
identically for all of them, so differences between rows in
Table~\ref{tab:main} reflect the models and not differences in metric
implementation. Each model is trained separately for each of the two feature
configurations, giving the twelve runs reported in the body. That routine is
\texttt{src/evaluation/unified\_eval.py} in the released code.

ConvLSTM and UTAE receive a five-day window, days $t{-}4$ through $t$, which is
the setting both were designed for and the reason they are included. The
remaining four receive day $t$ alone. UTAE also receives the day-of-year of each
observation, which its temporal attention encoder uses as a positional signal.
The four baselines released with the dataset are optimized with AdamW at its
default parameters, $\beta_1 = 0.9$, $\beta_2 = 0.999$ and $\lambda = 0.01$
\citep{gerard2023}, and we keep those settings. The released sweep
configurations train for 10k steps, except logistic regression at 520.
Checkpoints are written on best validation loss, but the metrics in
Table~\ref{tab:main} come from the unified evaluation pass at the end of each
run, so they reflect the final weights rather than the selected checkpoint.

\begin{table}[h]
\centering\small
\caption{Model and training configuration. Day $t$ is the most recent observed
day of a fire, and every model predicts the active fire mask on day $t{+}1$.
Parameter counts are for the \emph{All} configuration. The four dataset
baselines use the configurations released with the benchmark.}
\label{tab:appendix-training}
\vspace{4pt}
\begin{tabular}{llccll}
\toprule
Model & Family & Days & Params & Loss & LR\\
\midrule
Logistic Regression & Discriminative & 1 & 361   & Dice            & 0.1\\
ResNet18 U-Net      & Discriminative & 1 & 14.4M & Dice            & 0.001\\
BCE U-Net           & Discriminative & 1 & 32.5M & Weighted BCE    & $2\times10^{-4}$\\
ConvLSTM            & Discriminative & 5 & 240K  & Jaccard         & 0.01\\
UTAE                & Discriminative & 5 & 1.1M  & Weighted BCE     & 0.01\\
Flow Matching       & Generative     & 1 & 94.5M & MSE on velocity & $5\times10^{-5}$\\
\bottomrule
\end{tabular}
\end{table}

\textbf{Discriminative baselines.} Logistic Regression is a single $3 \times 3$
convolution applied to the input channels, following the released
implementation, and it is the floor for what the channel set alone supports.
ResNet18 U-Net is the segmentation-models-pytorch encoder--decoder released with
the dataset \citep{gerard2023,ronneberger2015,he2016}. ConvLSTM
\citep{shi2015} replaces the fully connected gates of an LSTM with convolutions,
so day-to-day dependence is modeled recurrently while spatial structure is
preserved, and it is used here as a single block followed by a convolution
producing the segmentation map. UTAE \citep{garnot2021} is a U-Net whose encoder
applies simplified multi-head self-attention across the temporal dimension at
the bottleneck, with the resulting attention map upscaled and applied to every
skip connection.

\textbf{Positive-class weighting.} Two of the six models optimize a binary
cross-entropy with an explicit positive-class weight, and they are the two
recall-maximizers. The released training script sets the UTAE fire-class weight
to the inverse relative frequency of that class in the training years,
overriding the value written in the configuration file \citep{wsts_repo}. At the
0.104\% base rate of 2018 and 2019 that gives 964. Our BCE U-Net uses a fixed
positive weight of 50, more than an order of magnitude smaller, and the two
models sit in that same order in Figure~\ref{fig:main}(b), with UTAE further
from the diagonal than BCE U-Net at precision 0.042 to 0.061 against 0.186 to
0.203. The three remaining discriminative baselines use unweighted overlap
losses, Dice for ResNet18 U-Net and Logistic Regression and Jaccard for
ConvLSTM, and all three sit near the diagonal. The size of the weight orders the
operating points, not the architecture family. For a weighted BCE loss with positive weight $w$, the boundary that an unweighted loss would place at 0.5 sits at $w/(1+w)$ on the raw score in expectation, which is 0.980 for BCE U-Net and 0.999 for UTAE. Thresholding at 0.5 therefore flags every pixel whose estimated fire probability exceeds about 2\% for BCE U-Net and about 0.1\% for UTAE, the latter being the training base rate.

\textbf{BCE U-Net.} A segmentation U-Net with 96 base channels, three
downsampling levels and multi-scale injection of the conditioning input at every
level, trained with binary cross-entropy under a positive weight of 50. The
model is optimized with AdamW at learning rate $2\times10^{-4}$, weight decay
$10^{-4}$, batch size 16, gradient clipping at 1.0 and a cosine annealing
schedule. A positive weight shifts the decision boundary away from 0.5 by
construction, so recall of 0.86 to 0.90 at precision of 0.19 to 0.20 is the
expected direction of effect. Section~\ref{sec:discussion} does not claim the
behavior is surprising. Its claim is that AP does not reveal it, since exact AP
depends only on the ranking of pixel scores. Our estimate is computed on a fixed
threshold grid, described in Appendix~\ref{app:eval}, and is therefore also
mildly sensitive to the scale on which those scores are expressed.

\textbf{Flow Matching.} Flow matching \citep{lipman2023,liu2023} learns a
continuous-time velocity field carrying a simple prior to the data distribution
and integrates it as an ODE at inference. Applied directly to a binary next-day
fire mask it fails at this sparsity, for two reasons. Integrating from the
current fire mask makes the regression target the displacement between
consecutive masks, which is overwhelmingly positive wherever new fire appears.
The learned field is therefore biased toward growth by construction, and
integration amplifies that bias into large connected over-predictions. A binary
mask is also a poorly posed regression target for a continuous field, since it
is almost entirely zeros with sparse unit spikes.

We therefore adopt the FlowSDF formulation
\citep{bogensperger2025flowsdfflowmatchingmedical}. Integration starts from
Gaussian noise instead of from the current mask, and the conditioning enters
through the network instead of through the integration start, which removes the
positive bias structurally. The regression target is the truncated signed
distance function of the day $t{+}1$ mask, negative inside the fire, positive
outside and zero at the boundary, truncated at 3 pixels. The distance field is
dense and smooth, so the loss carries signal everywhere and the model has to use
the conditioning. The loss is a plain mean squared error on the velocity field
with no asymmetric weighting, which the balanced target makes unnecessary. It is
the only model in Table~\ref{tab:main} that under-predicts.

Signed distance values are standardized once using statistics estimated over the
training masks and applied identically at training and test time, with mean
2.9699 and standard deviation 0.3237. The zero level set of the raw field, which
is the mask boundary, therefore maps to the fixed normalized value
$-\text{mean}/\text{std} = -9.175$, so mask recovery is deterministic. Because
the distance distribution is right-skewed at this sparsity, a small positive
residual skew of about 0.1 survives standardization, which the network learns
around. The recovered field is mapped to a bounded per-pixel score in $[0,1]$ by
a monotone transform that sends a raw signed distance of zero to exactly 0.5,
so the threshold metrics are computed at the model's own mask boundary and AP
is computed on a continuous score for all six models. That score occupies a narrower band than the sigmoid outputs of the
other five, which the fixed-grid AP estimate of Appendix~\ref{app:eval} is not
fully invariant to.

The velocity field is a conditional network with 128 base channels and a
256-dimensional time embedding, trained for 100 epochs at batch size 16 with
learning rate $5\times10^{-5}$, 500 warmup steps, weight decay $10^{-4}$, and
gradient clipping at 0.5. Inference integrates 50 Euler steps from noise, and
because the starting point is sampled, a single forward pass gives one draw from
the model and not a deterministic map. The row in Table~\ref{tab:main} is one
such draw, taken without a fixed random seed. Training also skips the optimizer
step on any batch whose loss is non-finite or more than five times a running
mean, which guards against a single pathological batch corrupting the weights.

\textbf{Protocol for the two custom models.} BCE U-Net and Flow Matching use the
same year split and the same preprocessing as the baselines, but they are
trained without a validation set, so no model selection was performed for them
at all. The periodic
evaluation used to monitor them during development ran on the 2021 test set, so
their formulation and hyperparameters were chosen with test metrics visible.
Their rows are best read as an upper bound on what each approach reaches on this
split rather than as a clean held-out estimate.

\textbf{Persistence baseline.} Carrying the day $t$ active fire mask forward
unchanged as the day $t{+}1$ prediction is the persistence forecast, a standard
reference baseline in short-range weather forecasting. It takes no parameters and reads no
channels beyond the fire mask, so it is identical under both feature
configurations. On the 2021 test year it reaches AP 0.287 and F1 0.535, against
a four-year mean of AP 0.193 and F1 0.432 \citep{gerard2023}. Every
architecture in Table~\ref{tab:main} exceeds it on AP, but only ResNet18 U-Net,
ConvLSTM and Logistic Regression exceed it on F1. BCE U-Net, UTAE and Flow
Matching all fall below a parameter-free baseline once a threshold is applied,
in both feature configurations, which is a second reading of the same split the
body draws between ranking quality and thresholded output.

\section{Evaluation Protocol}\label{app:eval}

\textbf{Metrics.} Precision is the fraction of predicted fire pixels that carry an active-fire detection on day~$t{+}1$, recall the fraction of day~$t{+}1$ active-fire pixels that were predicted, F1 their harmonic mean, and IoU the intersection over union of predicted and observed fire pixels. All four are computed at threshold 0.5 on the per-pixel score.
Average Precision summarizes the precision--recall curve, estimated on a fixed
grid of 200 uniformly spaced thresholds so that memory stays bounded over the
full test season. That estimate is close to exact AP for scores spread across
$[0,1]$ and drifts low for scores concentrated in a narrow band, so it should be
read as an approximation rather than as the exact statistic. Scores and targets are pooled across the
whole 2021 test set before the metrics are computed, not averaged per fire, so
large fires contribute in proportion to their pixel count.

\textbf{Redundancy between F1 and IoU.} Two of the reported metrics are
algebraically determined by the others. For a binary task,
$\text{IoU} = \text{F1}/(2 - \text{F1})$ exactly. This holds across all twelve
runs in Table~\ref{tab:main} to within 0.001, which is rounding. The IoU
columns therefore corroborate nothing the F1 columns do not already establish,
and the two are best read as one measurement and not as two that agree. We
report both because both are community standards in this domain, not because
they are independent evidence.

\textbf{Predicted-to-observed active-fire area.} The ratio of predicted to observed active-fire area is $R/P$, since predicted positives are $\mathrm{TP}/P$ and actual positives are $\mathrm{TP}/R$. Forecast verification calls this ratio the frequency bias, and IoU the critical success index \citep{roebber2009visualizing}. It is
recoverable from Table~\ref{tab:main} without further measurement, and it is the
quantity behind the area figures quoted in the abstract and the conclusion.
Table~\ref{tab:appendix-area} gives it for all twelve runs.

\begin{table}[h]
\centering\small
\caption{Predicted active-fire area divided by observed active-fire area, equal to $R/P$, at threshold 0.5 over the 2021 test year. A value of 1 matches the observed extent, above 1  over-predicts and below 1 under-predicts.}
\label{tab:appendix-area}
\vspace{4pt}
\begin{tabular}{lcc}
\toprule
Model & \emph{Vegetation} & \emph{All}\\
\midrule
BCE U-Net           & 4.82  & 4.24\\
UTAE                & 15.88 & 22.95\\
ResNet18 U-Net      & 1.04  & 0.96\\
ConvLSTM            & 1.07  & 1.10\\
Logistic Regression & 1.21  & 1.05\\
Flow Matching       & 0.82  & 0.84\\
\bottomrule
\end{tabular}
\end{table}

The three operating-point profiles separate cleanly here.
The recall-maximizers sit between 4.24 and 22.95, the balanced models within
0.96 to 1.21, and Flow Matching below 1.00 in both configurations. Flow Matching
is the only model under 1.00 in both, which is the basis for calling it the one under-predictor. ResNet18 U-Net falls marginally below on \emph{All} at 0.96 while
sitting above on \emph{Vegetation}, so it does not hold that profile across
configurations and is grouped with the balanced models throughout.

AP separates ConvLSTM and UTAE on \emph{Vegetation} by 0.403 against 0.383,
about five percent, while the area the two flag differs by a factor of roughly
fifteen. Exact AP is invariant to monotone rescaling of the scores, so a difference of that size lies outside what it can express by construction. The difference is a property of the operating point and not of the ranking, and a higher threshold for UTAE would shrink it.

\textbf{Thresholds in Figure~\ref{fig:qual}.} Figure~\ref{fig:qual} is rendered
at threshold 0.9 rather than the 0.5 used in Table~\ref{tab:main}, so that the
BCE U-Net panel stays legible, and it shows a single 2021 test fire rather than
the whole test year. Its 2.1$\times$ and the 4.82$\times$ in
Table~\ref{tab:appendix-area} are therefore not the same measurement. The figure is the more conservative of the two, since raising the threshold shrinks the predicted region, although 0.9 is still below the 0.980 boundary that the positive weight implies (Appendix~B).

\textbf{Comparison to the published baselines.} Our harness reproduces the
persistence baseline for 2021 exactly, at AP 0.287 against the published 0.287
\citep{gerard2023}. Because persistence is parameter-free, that match tests the
data loading and the target construction without any training in the way. It
does not exercise the continuous-score path of the AP estimator, since
persistence emits a binary score. Table~\ref{tab:appendix-published} sets the four shared
architectures against their published values.

\begin{table}[h]
\centering\small
\caption{AP for the four architectures shared with \citet{gerard2023}.
Published values are means over the full twelve-fold cross-validation with the
standard deviation across folds. Ours are fold 0, with 2021 held out.}
\label{tab:appendix-published}
\vspace{4pt}
\begin{tabular}{lcccc}
\toprule
& \multicolumn{2}{c}{\emph{Vegetation}} & \multicolumn{2}{c}{\emph{All}}\\
\cmidrule(lr){2-3}\cmidrule(lr){4-5}
Model & Published & Ours & Published & Ours\\
\midrule
Logistic Regression & 0.279 $\pm$ 0.092 & 0.352 & 0.286 $\pm$ 0.092 & 0.371\\
ResNet18 U-Net      & 0.328 $\pm$ 0.090 & 0.375 & 0.341 $\pm$ 0.086 & 0.401\\
ConvLSTM            & 0.306 $\pm$ 0.082 & 0.403 & 0.292 $\pm$ 0.094 & 0.390\\
UTAE                & 0.372 $\pm$ 0.088 & 0.383 & 0.321 $\pm$ 0.135 & 0.322\\
\bottomrule
\end{tabular}
\end{table}

All eight differences are positive and none exceeds 1.2 standard deviations of
the published spread across folds. Two documented effects explain the direction.
The first is that 2021 is the easiest of the four years by the dataset's own
measure, since persistence scores AP 0.287 there against a four-year mean of
0.193, so any model tested on 2021 alone should sit above a twelve-fold average
\citep{gerard2023}. The second is the corrected dataset class, which the
authors report gives slightly higher performance than the version the published
numbers were produced with \citep{wsts_repo}. We report
the comparison as a check on the harness rather than as a like-for-like result,
since one fold and twelve folds are not the same experiment.

\end{document}